\documentclass[10pt,twocolumn,letterpaper]{article}

\usepackage[pagenumbers]{cvpr} 
\usepackage{comment}

\usepackage[dvipsnames]{xcolor}

\definecolor{glaucous}{rgb}{0.38, 0.51, 0.71}
\definecolor{stop_grad_red}{rgb}{0.75, 0, 0}

\definecolor{cvprblue}{rgb}{0.21,0.49,0.74}
\usepackage[pagebackref,breaklinks,colorlinks,citecolor=cvprblue]{hyperref}

\def\paperID{527} 
\def\confName{CVPR}
\def\confYear{2024}

\title{High-level Feature Guided Decoding for Semantic Segmentation}
\author{Ye Huang\textsuperscript{\rm 1}\quad
        Di Kang \textsuperscript{\rm 2}\quad
        Shenghua Gao \textsuperscript{\rm 3}\quad
        Wen Li \textsuperscript{\rm 1}\quad
        Lixin Duan\textsuperscript{\rm 1}\thanks{Corresponding author}\\
{\small \textsuperscript{\rm 1} University of Electronic Science and Technology of China}\\
{\small
    \textsuperscript{\rm 2} Tencent AI Lab \quad
    \textsuperscript{\rm 3} ShanghaiTech University
}
}

\begin{document}

\twocolumn[{%
\renewcommand\twocolumn[1][]{#1}%
\maketitle
\begin{center}
\centering
\includegraphics[width=\textwidth]{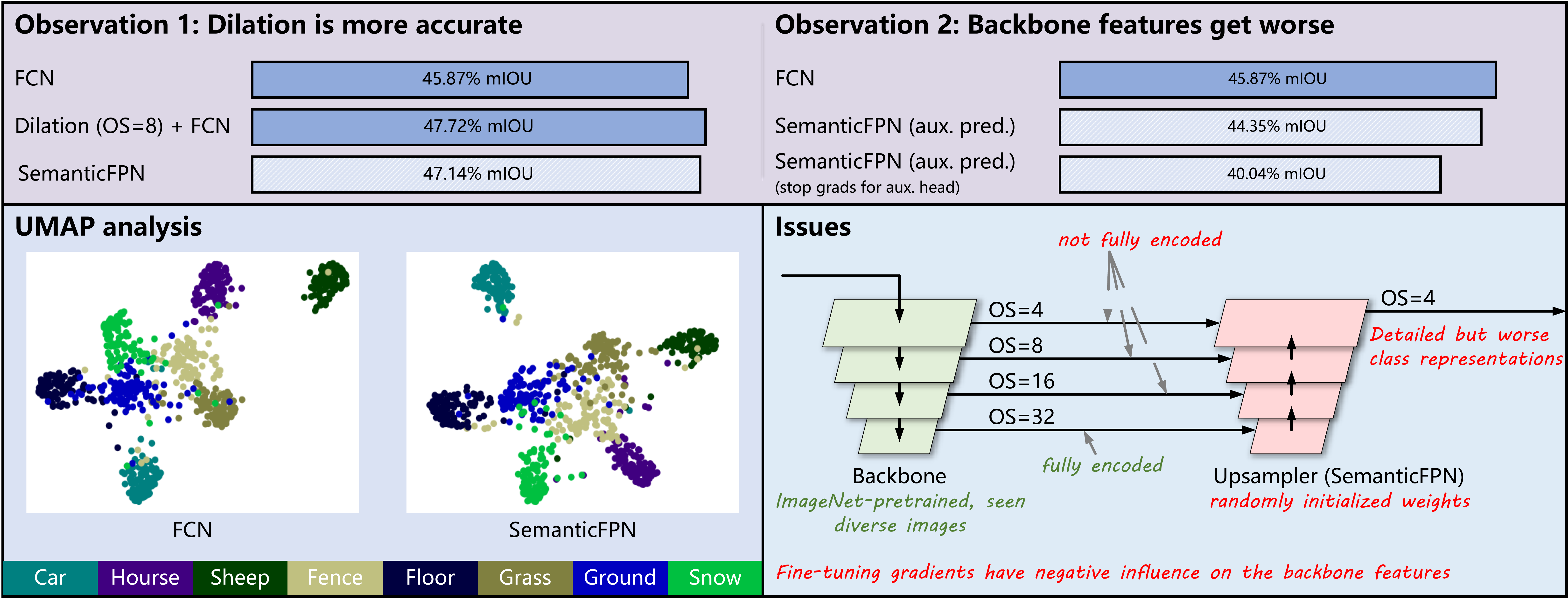}
\captionof{figure}{
\textbf{Problem identification.}
We notice that 1) using an upsampler net (i.e. SemanticFPN) is worse (\eg, more entangled) than using a dilated backbone; 2) the joint fine-tuning of the upsampler and the backbone usually results in worse backbone features (see Sec.~\ref{sec:problem}).
In this work, we propose to use high-quality \textit{\textbf{h}}igh-level \textit{\textbf{f}}eatures to \textit{\textbf{g}}uide the training of the upsampler.
}
\label{fig:hfgd:concept}
\end{center}}]

\begin{abstract}
\noindent Existing pyramid-based upsamplers (\eg SemanticFPN), although efficient, usually produce less accurate results compared to dilation-based models when using the same backbone. 
This is partially caused by the \emph{contaminated} high-level features since they are fused and fine-tuned with noisy low-level features on limited data.
To address this issue, we propose to use powerful pre-trained \textit{\textbf{h}}igh-level \textit{\textbf{f}}eatures as \textit{\textbf{g}}uidance (HFG) 
so that the upsampler can produce robust results.
%
%
Specifically, \emph{only} the high-level features from the backbone are used to train the class tokens, which are then reused by the upsampler for classification, guiding the upsampler features to more discriminative backbone features.
One crucial design of the HFG is to protect the high-level features from being contaminated by using proper stop-gradient operations so that the backbone does not update according to the noisy gradient from the upsampler.
To push the upper limit of HFG, we introduce a \textit{\textbf{c}}ontext \textit{\textbf{a}}ugmentation \textit{\textbf{e}}ncoder (CAE) that can efficiently and effectively operate on the low-resolution high-level feature, resulting in improved representation and thus better guidance.
We named our complete solution as the High-Level Features Guided Decoder (HFGD).
We evaluate the proposed HFGD on three benchmarks: Pascal Context, COCOStuff164k, and Cityscapes.
HFGD achieves state-of-the-art results 
among methods that do not use extra training data,
demonstrating its effectiveness and generalization ability. 
\end{abstract}    
\section{Introduction}
\label{sec:hfgd:intro}

Semantic Segmentation is a fundamental task in computer vision, which densely predicts per-pixel class for a given image.
Most existing methods include an encoder (\ie{~backbone}) to output coarse (low-resolution) high-level features and a decoder to recover lost spatial information to output high-resolution final results.

Due to the lack of densely labeled images, existing segmentation methods typically use pre-trained backbones on large-scale image classification datasets to extract high-level semantic features that are more expressive and generalizable\footnote{The scales of common semantic segmentation datasets vary from 2.9K training images in Cityscapes to 118K in COCOStuff while ImageNet-1K/21K contains 1.2M/14M images.}.
Popular classification backbones  (\eg ResNet~\cite{cResnet}, EfficientNet~\cite{cEfficientNet}, Swin~\cite{cSwin} and ConvNeXt~\cite{cConvNeXT}) 
typically include multiple downsampling layers, resulting in small feature maps with an output stride~\cite{cDeepLab} of 32 (OS=32).
However, the feature maps generated by the OS=32 are too coarse to obtain per-pixel classifications that are accurate and detailed enough, such as object boundaries and small/thin objects.
So, many recent works have investigated stronger and more efficient upsamplers.

In one line of works, dilation-based methods~\cite{cDeepLab,cPSPNet,cDeepLabV3Plus,cCCNet,cCFNet,cACFNet,cOCR,cCPN,cDualAttention,cDenseASPP} replace the downsampling operations in the backbone with dilated convolutions to prevent resolution reduction.
Dilation-based methods converge faster and generalize better with effective utilization of pre-trained weights and no introduction of random weights.
The limitations of dilation are obvious: it only applies to CNN-based backbones and comes with high computational cost.

In another line of works~\cite{cUper,cFaPN}, an upsampler branch is introduced (\eg SemanticFPN~\cite{cPanopticFPN}) to incorporate fine details from low-level features into upsampled high-level features.
Although computationally more efficient, they are often less effective than dilation-based methods since new random weights are required, and they can only be trained with limited segmentation data.
What's worse, the originally good high-level features may be contaminated during the fine-tuning (Fig.~\ref{fig:hfgd:concept} and Sec.~\ref{sec:problem}).

After identifying this root cause, our key insight is to \emph{protect} the good (high-level) backbone features and use its \textit{\textbf{h}}igh-level \textit{\textbf{f}}eatures as \textit{\textbf{g}}uidance (HFG) during the training of the upsampler branch.
Specifically, we first isolate/protect the backbone from the upsampler via proper stopping gradient operations.
Secondly, the training of the upsampler is guided/regularized by the good high-level backbone features since we force the upsampler features to mimic the backbone features.
From the perspective of distillation, the backbone branch is the teacher, and the upsampler branch is the student, except that the upsampler (student) uses information from the backbone (teacher) since they are parts of one single network.

To further push the upper limit of HFG, we propose to introduce an \textit{\textbf{c}}entext \textit{\textbf{a}}ugmented \textit{\textbf{e}}ncoder (CAE) to enhance the high-level backbone feature.
Experimentally, our CAE design can effectively operate on OS=32 feature maps, striking a good balance between efficiency and effectiveness.
We also make some modifications to SemanticFPN to better suit HFGM and (optionally) produce more detailed predictions in OS=2.

In summary, our contributions include:
\begin{itemize}
    \item We first identify the issues that limit the accuracy of existing pyramid-based upsamplers (\eg SemanticFPN).
    \item  
    We propose to use \textit{\textbf{h}}igh-level \textit{\textbf{f}}eatures as \textit{\textbf{g}}uidance (HFG) for the upsampler learning and protect the backbone features from being contaminated as previous methods.
    \item To further push the upper limit, we propose \textit{\textbf{c}}entext \textit{\textbf{a}}ugmented \textit{\textbf{e}}ncoder (CAE) to enhance the high-level guidance features and 
    modify SemanticFPN to fit the proposed HFG better.
    \item 
    Our full HFG decoder(HFGD) achieves state-of-the-art accuracy on Pascal Context, COCOStuff164k, and Cityscapes test set among methods that do not use extra training data.
\end{itemize}
\section{Related Works}
\label{sec:HFGD:related_work}

\noindent\textbf{Dilated convolutions.} 
Dilated CNNs~\cite{cDeepLabV3,cDeepLabV3Plus,cMSCA} usually use dilated convolution layers starting from $OS=8$ feature maps, resulting in the constant resolution thereafter.
Dilation models are easy and fast to train since only a few randomly initialized weights are appended to the well-trained high-level backbone features.
However, the computation cost for the final layers increases quadratically with the length of the feature map.
Another issue is that dilation is only applicable for CNN backbones (\eg VGG~\cite{cVGG}, ResNet~\cite{cResnet}, and ConvNeXt~\cite{cConvNeXT}) but not for Transformer-based backbones (\eg ViT~\cite{cViT} and Swin~\cite{cSwin}).
 
\noindent\textbf{Pyramid-based upsampler.}
Pyramid-based upsampler is another popular strategy, such as UNet~\cite{cUNet}, SemanticFPN~\cite{cFPN,cPanopticSeg}, Uper~\cite{cUper} and FaPN~\cite{cFaPN}.
They often progressively merge features from different stages of the backbone network (\ie a feature pyramid) into the upsampled high-level feature maps from $OS=32$, resulting in much less computation cost than using a dilated backbone.

However, their accuracy is usually inferior to dilation, which is implicitly mentioned in many works. 
For example, DeepLab V3+ still uses the dilated backbone in the final; UperNet has a slightly worse mIOU than PSPnet, even with more advanced training strategies such as Synced Batch Norm. 
One reason is that they introduce more randomly initialized weights, especially those weights used to process early-stage low-level features.
Its drawbacks are two-fold.
Firstly, the new learnable weights are trained only with limited semantic segmentation data.
Secondly, directly updating the early-stage backbone features through a deep-supervision/shortcut fashion with such small amount of data will probably be harmful for generalization (Fig.~\ref{fig:hfgd:concept}).

\begin{figure*}[th]
\centering
\includegraphics[width=1.0\linewidth]{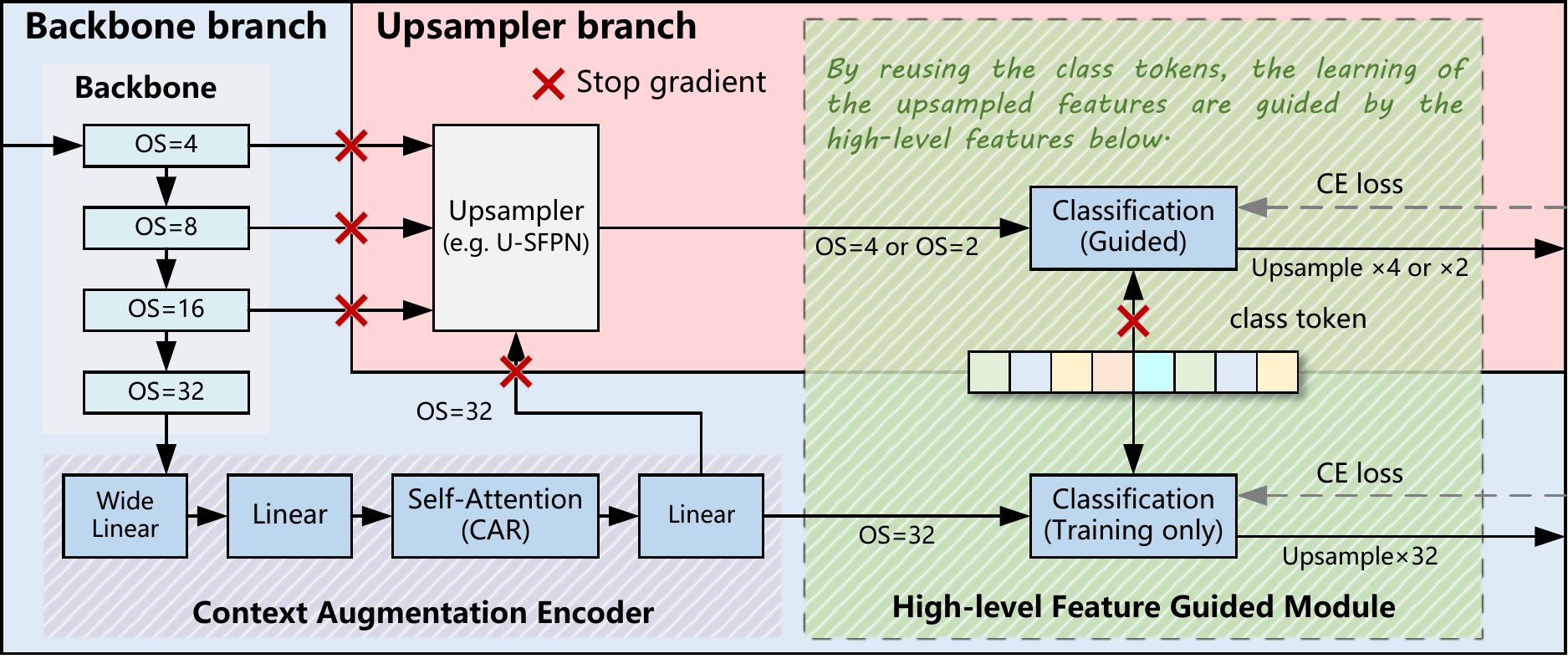}
\caption{
\textbf{High-level Feature Guided Decoder (HFGD).}
The core of HFGD is the high-level feature guided module that guides the upsampler learning and protects pre-trained backbone features by properly stopping gradients.
Then a context augmentation encoder (CAE) is included after the backbone to push the upper limit of HFG efficiently and effectively.
Lastly, a slightly modified ultra SemanticFPN (U-SFPN) is proposed to fit our HFGM/CAE better and supports operating on even higher resolution (OS=2).
Refer to Sec.~\ref{sec:method} for details.
}
\label{fig:URD:Arch}
\end{figure*}

\noindent\textbf{Context encoding modules.} 
There exist many powerful modules focusing on context encoding, such as pyramid pooling module (PPM)~\cite{cPSPNet}, 
atrous spatial pyramid pooling (ASPP)~\cite{cDeepLab,cDeepLabV3,cDeepLabV3Plus}, 
Self-Attention~\cite{cNonLocal,cDualAttention,cAttentionIsAllYourNeed}, etc.
They all effectively enlarge the receptive field size and aggregate more context information to obtain more powerful pixel representations.
However, their computational costs are often huge and increase quadratically with the length of the feature map.
To strike a good balance between efficiency and effectiveness, we insert the extra encoding module after the OS=32 feature maps.
Under this setting, we found them less effective or harmful, possibly because they were originally proposed for processing OS=8 feature maps.
Finally, we make some modifications on a CAR-based~\cite{cCAR} so that it suits better to the HFG.

\section{Proposed Method}
\label{sec:method}

The key to our method is using \textit{\textbf{h}}igh-level \textit{\textbf{f}}eature as \textit{\textbf{g}}uidance (HFG) to effectively guide the upsampler training without introducing side-effect to the well-trained backbone branch (Sec.~\ref{sec:method:hfgm}).
Since the high-level features play a crucial role, we design a context-augmented encoder (CAE) that operates on OS=32 to strike a good balance between efficiency and effectiveness (Sec.~\ref{sec:method:cae}).
Lastly, we make some modifications based on SemanticFPN (SFPN)~\cite{cPanopticFPN} (Sec.~\ref{sec:method:upsampling}) so that it better fits our HFGM and CAE.

\subsection{High-level feature guided module (HFGM)}
\label{sec:method:hfgm}

\noindent\textbf{Motivation.} 
In our high-level feature guidance module,
we treat the high-level feature branch as the teacher (\ie not updated by the gradient from the student) and the low-level feature branch as the student.

The last stage high-level features (OS=32) extracted from a classification backbone, are already ``good'' representations for semantic segmentation since it is essentially a per-pixel classification task.
However, the high-level features are too coarse ($OS=32$) for accurate segmentation.
In contrast, features from the earlier stages are in higher resolution and may contain some useful details ignored by the aforementioned high-level features.
%
However, a substantial gap exists between low-level backbone features and the final segmentation feature.
Thus, a calibration, guided by the high-level features, on the low-level features is required to suit the final segmentation task better.

\noindent\textbf{Guiding the upsampler.} 
Guidance is realized by reusing the class tokens and a stop-gradient operation.
Specifically, the backbone and upsampler branches share the same set of class tokens to calculate class probabilities and the cross-entropy loss.
The class tokens are trained to co-evolve with only the high-level features (\ie the guiding branch). 
They are not updated by the gradients from the upsampler branch via the stop-gradient operations (\ie the guided branch).
As a result,  the upsampler branch pixels are forced to approach their corresponding high-level pixel representations from the backbone branch to improve generalization. 

\noindent\textbf{Protecting pre-trained backbone.} 
It is important to protect the backbone from being contaminated by the training of the upsampler, as evidenced by observation 2 in Fig.~\ref{fig:hfgd:concept}.
We realize this goal with proper stop-gradient operations between the upsampler and backbone branches, resulting in a teacher-student~\cite{cBYOT} like distillation architecture (Fig.~\ref{fig:URD:Arch}).

Specifically, the stop-gradient operations on $OS=n$ paths (between the upsampler branch and the backbone branch) stop updates from randomly initialized upsampler branch so that they can protect well-trained backbone features from being negatively affected.

Another small but beneficial operation is the inclusion of an axial attention~\cite{cAxialAttention,cAxialDeepLab,cCAA} layer.
This axial attention broadcasts the gradient of one pixel to all spatial locations, resulting in further improvement (see Tab.~\ref{tab:ablation_hfgm}).
In consideration of computation, we choose axial attention~\cite{cAxialAttention,cAxialDeepLab,cCAA} instead of full self-attention for spatial context augmentation since the resolution of the path is high (\eg $OS=2$ or $4$).
Note that axial attention acts as a helper for the guidance, possibly because it effectively broadcasts the guidance signal back to all spatial locations.
Without guidance, its effect is limited (Tab.~\ref{tab:ablation_hfgm})

\subsection{Context-augmented encoder (CAE)}
\label{sec:method:cae}

To further push the upper limit, we propose to enhance the high-level features that serve as the guidance/teacher.
Considering computation overhead, we choose to operate on the low-resolution OS=32 feature maps to use any powerful modules (\eg ASPP~\cite{cDeepLab,cDeepLabV3,cDeepLabV3Plus}, PPM~\cite{cPSPNet}, OCR~\cite{cOCR}, and full self-attention~\cite{cNonLocal,cDualAttention,cAttentionIsAllYourNeed,cCFNet}) freely.

In our experiments, we propose a CAR-based~\cite{cCAR} configuration that achieves most accuracy gain (see Tab.~\ref{tab:ablation_cae}).
Specifically, we use a wider linear layer (\ie $1\times1$ conv with 2048 channels) to process backbone features.
Then, we use a linear layer to reduce the features to 512-D, followed by a full self-attention layer regularized by class-aware regularizations (CAR\cite{cCAR}) and a trailing convolution layer (256-d).
Using a wider linear layer (\ie wide linear in Fig.~\ref{fig:URD:Arch}) performs much better than the commonly used $3\times3$ conv layer, possibly because the receptive field size of a pixel in OS=32 feature maps is already large enough.
Increasing its channel number also helps since it compensates for the lost capacity after changing to smaller kernels.

\subsection{Ultra SemanticFPN (U-SFPN)}
\label{sec:method:upsampling}
We slightly modify SemanticFPN (SFPN) to fit our HFGM and CAE better and enable it to work on even higher-resolution (OS=2) feature maps.
Firstly, we change the leading $1\times1$ conv to $3\times3$ conv layers since it collects more spatial information to reduce potential ``noise`` in the low-level features.
Secondly, we increase the channel numbers from 128 to 256 (applied to SemanticFPN in experiments for fairness) since we find it beneficial to obtain more accurate and detailed results.
Last but not least,
we can keep upsampling to an even higher resolution (\ie $OS=2$) and achieve better accuracy with the help of HFG\footnote{We only test $OS=2$ results on Cityscapes since the ground truth annotations on other datasets (\eg Pascal Context) are not as accurate.}.
To achieve this goal, we include one more ``conv-upsample'' block for every branch to get $OS=2$ feature maps.
Note that we did not use $OS=2$ low-level features from the backbone.

\section{Training details}
\label{sec:HFGD:training_settings}

Our training settings, unless specified, strictly follow CAR~\cite{cCAR}.
When performing ablation studies, we use the plain setting, including SGD optimizer and learning, to provide a simple, \textbf{clean setting} to study method effectiveness.
When comparing with state-of-the-arts, we utilize the advanced settings. 
We applied training settings as follows:
\begin{table}[h]
\centering
\small
\resizebox{\linewidth}{!}{%
\begin{tabular}{l|c|c} 
    \toprule
    Settings    & Plain~\cite{cCCNet,cCAR} \quad & Advanced~\cite{cCAR} \quad \\
    \midrule
    \midrule
    Batch size & 16 & 16  \\
    Optimizer & SGD & AdamW \\
    Learning rate decay & \textit{poly}  & \textit{poly} \\
    Initial Learning rate & 0.01 & 0.00004 \\
    Weight decay & 0.0001  & 0.05 \\
    Photo Metric Distortion & - & \checkmark \\
    Sync Batch Norm & \checkmark & \checkmark \\
    \bottomrule
\end{tabular}
}
\label{tab:urd:ablation_training_settings}
\end{table}
\section{Experiments on Pascal Context Dataset}
The Pascal Context~\cite{cPascalContext} dataset contains 4,998 training images and 5,105 testing images.
Following the common practice, we use its 59 semantic classes to conduct the ablation studies and experiments.
Unless specified, we train the models on the training set for 30K iterations for the ResNet backbone and 40K for Swin-Large and ConvNeXt-Large.


\begin{table}[t]
\centering
\small
\resizebox{\linewidth}{!}{%
\begin{tabular}{c|l|c|c} 
\toprule
Backbone & Dilated or Upsampler & Extra Encoding & mIOU(\%)  \\
\midrule
\midrule
ResNet-50 & -  &  Identity      &  45.87 \\
ResNet-50 & Dilation, $OS=8$ &  Identity      & \textbf{47.72}  \\
ResNet-50 & SFPN, $OS=4$ &  Identity      & 47.14  \\
\midrule
ResNet-50 & -  & ASPP~\cite{cDeepLabV3Plus}           & 46.41  \\
ResNet-50 & Dilation, $OS=8$  & ASPP~\cite{cDeepLabV3Plus}           & \textbf{48.59}  \\
ResNet-50 & SFPN, $OS=4$ & ASPP~\cite{cDeepLabV3Plus}           &  47.81 \\
\midrule
ResNet-50 & -  & OCR~\cite{cOCR}            & 45.50 \\
ResNet-50 & Dilation, $OS=8$ & OCR~\cite{cOCR}            & \textbf{48.23} \\
ResNet-50 & SFPN, $OS=4$ & OCR~\cite{cOCR}            & 47.39 \\
\midrule
ResNet-50 & - & SA~\cite{cNonLocal} & 45.02 \\
ResNet-50 & Dilation, $OS=8$ & SA~\cite{cNonLocal} & \textbf{48.32} \\
ResNet-50 & SFPN, $OS=4$ & SA~\cite{cNonLocal} & 45.90 \\
\midrule
ResNet-50 & - & SA~(CAR~\cite{cCAR}) &  47.18  \\
ResNet-50 & Dilation, $OS=8$ & SA~(CAR~\cite{cCAR}) & \textbf{50.50} \\
ResNet-50 & SFPN, $OS=4$ & SA~(CAR~\cite{cCAR}) & 48.51  \\
\bottomrule
\end{tabular}
}
\caption{
Comparisons between a dilation method and a state-of-the-art upsampler-based method (i.e. SemanticFPN) with various extra encoding techniques on the Pascal Context dataset.
Results demonstrate that dilation is more accurate than SemanticFPN with a substantial margin.
Identity means using no extra encoding (i.e. a basic FCN).
}
\label{tab:ablation_dilation_vs_fpn}
\end{table}

\subsection{Problem verification}
\label{sec:problem}

\noindent\textbf{SemanticFPN is generally worse than dilation.}
We conduct experiments in Tab.~\ref{tab:ablation_dilation_vs_fpn} to demonstrate the issue we identify (observation 1 in Fig.~\ref{fig:hfgd:concept}).
Many famous methods for context encoding (``Extra Encoding'') produce less accurate predictions when combined with an upsampler net (i.e. SemanticFPN) than directly using a dilated backbone, even if the upsampler produces higher-resolution feature maps.
Refer to our appendix, for detailed settings.

\noindent\textbf{Negative influence from the upsampler on the backbone.}
We find the joint fine-tuning of the upsampler and the backbone results in deteriorated backbone features.
To demonstrate it, we modify SemanticFPN by introducing an auxiliary FCN to predict the mask from the high-level features produced by the backbone (see Issues in Fig.~\ref{fig:hfgd:concept}).
The predictions from the auxiliary FCN branch become worse than the original FCN (44.35\% vs 45.87\% mIOU).
Another variant stops the auxiliary FCN's gradient from propagating back to the backbone, which produces even worse results (40.04\%).

\noindent\textbf{Advantage of pre-training.}
We conduct experiments to show the benefit of using the pre-trained backbone in Tab.~\ref{tab:ablation_imagenet_simple}.
Due to limited training data, there still exists a substantial gap after 6 times training iterations.
Thus, protecting the backbone to ensure its generalization is necessary and critical.
More experiments about this are presented in the supplementary.


\subsection{Ablation studies on HFGD}

\noindent\textbf{Ablation studies on HFGM.}
In Tab.~\ref{tab:ablation_hfgm}, we evaluate the effectiveness of our proposed HFGM based on SemanticFPN (\ie no ``Extra Encoding'').
Although using only HFG (``+guidance'') or only axial attention (``+AA'') is helpful, using the full HFGM brings the most gain (1.80\% mIOU).
probably because AA can effectively broadcast the guidance information of HFG to all spatial locations (also see Sec.~\ref{sec:method:hfgm} for more discussion).

\noindent\textbf{Importance of stopping gradients.}
We use several stop-grad operations to protect the backbone weights, especially the early low-level weights, and only allow gradients from the backbone branch to update their weights gradually (see. Fig.~\ref{fig:URD:Arch}).
We tested removing all stop-grad operations and obtained significantly decreased accuracy (48.94\% vs 48.50\%).
%

\noindent\textbf{Ablation studies on CAE.}
%
%
In Tab.~\ref{tab:ablation_cae}, we compare with CAR under different settings on the Pascal Context dataset 
since CAR~\cite{cCAR} performs best in Tab.~\ref{tab:ablation_dilation_vs_fpn} and CAE is based on CAR~\cite{cCAR}.
Results show that CAE is more compatible than CAR when using SFPN and ``SFPN + HFGM''.
Combined with results in Tab.~\ref{tab:ablation_dilation_vs_fpn}, our CAE design outperforms all the other alternatives (``Extra Encodings'') with and without SemanticFPN and approaches the accuracy of the best dilation-based model in Tab.~\ref{tab:ablation_dilation_vs_fpn}.

Similar to experiments in Tab.~\ref{tab:ablation_hfgm},
we also analyze the effects of CAE with different HFGM settings.
HFGM now provides better guidance to the upsampler with the help of CAE, leading to further improved final results (50.28\% vs 48.94\% and 49.22\% vs 47.67\% in Tab.~\ref{tab:ablation_hfgm}).


\begin{table}[t]
\centering
\small
\resizebox{\linewidth}{!}
{\def\arraystretch{1} \tabcolsep=0.6em 
\begin{tabular}{l|c|c|c} 
\toprule
Training Iterations & 30K & 90K & 180K  \\
\midrule
\midrule
ResNet-50 (ImageNet) + FCN & 45.87 & - & - \\
ResNet-50 (scratch) + FCN  & 26.13 & 31.38 & 34.00 \\
\bottomrule
\end{tabular}
}
\caption{
Importance of using ImageNet pre-trained weights.
Experiments are conducted on the Pascal Context dataset (mIOU\%).
%
}
\label{tab:ablation_imagenet_simple}
\end{table}

\begin{table}[t]
\centering
\small
\resizebox{\linewidth}{!}
{
\begin{tabular}{c|c|c|c|l} 
\toprule
Backbone   & Upsampler  & Extra Encoding & HFGM & mIOU(\%)  \\
\midrule
\midrule
ResNet-50  & SFPN & Identity & -      & 47.14  \\
ResNet-50  & SFPN & Identity & + guidance & 47.67 (\textcolor{black}
{$+$0.53}) \\
ResNet-50  & SFPN & Identity & + AA  & 47.88 (\textcolor{black}
{$+$0.74})  \\
ResNet-50  & SFPN & Identity & full  & 48.94 (\textcolor{black}{$+$1.80)}\\
\bottomrule
\end{tabular}
}
\caption{
Ablation studies on different HFGM settings on Pascal Context dataset.
Using both guidance and AA in HFGM brings the most gain (1.8\% mIOU).
AA and the proposed high-level feature guidance cooperate well probably because AA can effectively back-propagate the guidance signal to all spatial locations.
}
\label{tab:ablation_hfgm}
\end{table}

\begin{table}[t]
\centering
\small
\resizebox{\linewidth}{!}{%
\begin{tabular}{l|c|c|c} 
\toprule
Methods & HFGM & Extra Encoding &mIOU(\%)  \\
\midrule
\midrule
ResNet-50  & &  SA (CAR) & 47.18  \\
ResNet-50  & & Our CAE& \textbf{47.35} \\
\midrule
ResNet-50 + SFPN & & SA (CAR) & 48.51  \\
ResNet-50 + SFPN  &&  Our CAE& \textbf{48.76} \\
\midrule
ResNet-50 + SFPN &\checkmark & SA (CAR) & 50.00  \\
ResNet-50 + SFPN & \checkmark & Our CAE& \textbf{50.28} \\
\bottomrule
\end{tabular}
}
\caption{
Ablation studies on CAE on Pascal Context dataset.
Results indicate our CAR-based CAE is more compatible than CAR~\cite{cCAR} with the ResNet-50 (OS=32), SFPN, and ``SFPN + HFGM''.
}
\label{tab:ablation_cae}
\end{table}
\begin{table}[t]
\centering
\small
\resizebox{\linewidth}{!}
{
\begin{tabular}{c|c|c|c|l} 
\toprule
Backbone   & Upsampler  & Extra Encoding & HFGM & mIOU(\%)  \\
\midrule
\midrule
ResNet-50  & SFPN & Our CAE & -      & 48.76    \\
ResNet-50  & SFPN & Our CAE & + guidance & 49.22 (\textcolor{black}{$+$0.46}) \\
ResNet-50  & SFPN & Our CAE & + AA & 49.06 (\textcolor{black}{$+$0.26}) \\
ResNet-50  & SFPN & Our CAE & full  & \textbf{50.28} (\textcolor{black}{$+$1.52}) \\
\bottomrule
\end{tabular}
}
\caption{
Ablation studies to analyze the effects of CAE with different HFGM settings on the Pascal Context dataset.
HFGM provides better guidance thanks to the improved high-level features extracted by CAE.
}
\label{tab:ablation_hfgm_cae}
\end{table}

\noindent\textbf{Ablation studies on U-SFPN.}
In Tab.~\ref{tab:ablation_upsampling}, we conduct ablation studies on U-SFPN to verify the effectiveness of our modification on SFPN while fixing CAE and HFGM.
Replacing SFPN with U-SFPN improves the mIOU of ResNet-50 (CAE + HFGM) by 0.48\%, reaching 50.76\%.
The improvement is even larger (1.14\%) when using Swin-Large as the backbone 
.

\begin{table}[t]
\centering
\small
\resizebox{\linewidth}{!}
{\def\arraystretch{1} \tabcolsep=0.6em 
\begin{tabular}{c|c|c|c|l} 
\toprule
Backbone & CAE & HFGM & Upsampler & mIOU(\%)  \\
\midrule
\midrule
ResNet-50  & \checkmark & \checkmark & SFPN & 50.28 \\
ResNet-50  & \checkmark & \checkmark & U-SFPN   & \textbf{50.76} (\textcolor{black}{$+$0.48}) \\
\midrule
Swin-Large & \checkmark & \checkmark & SFPN & 59.32 \\
Swin-Large & \checkmark & \checkmark & U-SFPN   & \textbf{60.46} (\textcolor{black}{$+$1.14}) \\
\bottomrule
\end{tabular}
}
\caption{Ablation studies on different upsampling heads on the Pascal Context dataset.
Our proposed U-SFPN upsampling head consistently outperforms semantic FPN (SFPN) for both CNN and Transformer backbones.
}
\label{tab:ablation_upsampling}
\end{table}

\noindent\textbf{Module-level ablation studies on HFGD.}
In Tab.~\ref{tab:ablation_hfgd},
%
We performed ablation studies on each module of HFGD using the best configurations found in Tab.~\ref{tab:ablation_cae}-\ref{tab:ablation_upsampling}. Using all modules together significantly improved accuracy, indicating the overall architecture's effectiveness.

\begin{table}[t]
\centering
\small
\resizebox{\linewidth}{!}
{\def\arraystretch{1} \tabcolsep=0.8em 
\begin{tabular}{c|c|l|c|c} 
\toprule
Backbone & Extra Encoding & Upsampler & HFGM & mIOU(\%)  \\
\midrule
\midrule
ResNet-50 & Identity & SFPN & - & 47.14 \\
ResNet-50 & Our CAE & SFPN    & -     & 48.76 \\
ResNet-50 & Identity  & U-SFPN & -     & 46.99\\
ResNet-50 & Our CAE& U-SFPN & - & 48.67 \\
ResNet-50 & Our CAE & U-SFPN & \checkmark & \textbf{50.76} \\
\midrule
Swin-Large & Our CAE& SFPN     &  -    & 56.78 \\
Swin-Large & Identity & U-SFPN &  - & 58.69 \\
Swin-Large & Our CAE & U-SFPN &  - & 55.76 \\
Swin-Large & Our CAE & U-SFPN & \checkmark & \textbf{60.46} \\
\bottomrule
\end{tabular}
}
\caption{Ablation studies on the proposed three modules using 
the best configurations found in Tab.~\ref{tab:ablation_cae}-\ref{tab:ablation_upsampling} on Pascal Context dataset.
}
\label{tab:ablation_hfgd}
\end{table}

\begin{table}[h]
\centering
\small
\resizebox{\linewidth}{!}
{\def\arraystretch{1} \tabcolsep=0.6em 
\begin{tabular}{l|c|c|c|l} 
\toprule
Backbone   & Upsampler  & EE & HFGM & mIOU(\%)  \\
\midrule
\midrule
R50  & U-SFPN & Our CAE & -     & 48.67   \\
R50  & U-SFPN & Our CAE & \checkmark  & \textbf{50.76} (\textcolor{black}{$+$2.09}) \\
\midrule
R50  & FaPN & - & -      &  47.50 \\
R50  & FaPN & - & \checkmark  &  \textbf{49.87} (\textcolor{black}{$+$2.37}) \\
\midrule
R50  & Uper & PPM & -      & 48.25 \\
R50  & Uper & PPM & \checkmark  & \textbf{49.96}   (\textcolor{black}{$+$1.71}) \\
\midrule
R50 (D8) & DeepLabV3+ & ASPP & -      &  48.11\\
R50 (D8) & DeepLabV3+ & ASPP & \checkmark  & \textbf{49.70} (\textcolor{black}{$+$1.59}) \\
\bottomrule
\end{tabular}
}
\caption{
Ablation studies on different upsamplers w/o or w/ our HFGM on Pascal Context dataset.
\textit{EE}: Extra encoding.
\textit{D8}: 
Dilated convolutions with ($OS=4$).
}
\label{tab:ablation_hfg_other_upsamplers}
\end{table}

\subsection{Computational cost of HFGD}
The computational cost of our HFGD and two other state-of-the-art methods are listed in Tab.~\ref{tab:HFGD:flops}.
HFGD uses much lower GFLOPs than a similar dilation model (Self-Attention + CAR~\cite{cCAR}) but achieves better mIOU (50.76\% vs 50.50\%~\cite{cCAR}).
Compared with SemanticFPN, HFGD ($OS=4$) achieves 3.62\% 

\begin{table}[t]
\centering
\footnotesize
\resizebox{\linewidth}{!}{%
\begin{tabular}{l|l|l|l}
\toprule
Method & Backbone & GFLOPs & mIOU\%\\
\midrule
\midrule
SA (CAR) & ResNet-50 (D8) & 158.96 & 50.50\\
\midrule
SemanticFPN & ResNet-50  & 45.65  & 47.14  \\
\midrule
HFGD (OS=4) &  ResNet-50 & 71.63  &  50.76 \\
HFGD (OS=2) &  ResNet-50 & 153.62 & 51.00  \\
\bottomrule
\end{tabular}
}
\caption{
Computation analysis of HFGD on $513\times513\times3$ images.
Although more efficient, previous upsampling-based methods (e.g. SemanticFPN) produce less accurate results than the dilation-based methods.
HFGD ($OS=4$) closes this accuracy gap while still being efficient.
Given a similar computation budget, HFGD ($OS=2$) further improves the accuracy (0.24\% mIOU).
}
\label{tab:HFGD:flops}
\end{table}

\subsection{Comparison with the state-of-the-art methods}

To compare with the state-of-the-art, we adopt ConvNeXt-L as the backbone for our HFGD.
We set the training iterations to 40K while
all the other training settings are the same as stated in Sec.~\ref{sec:HFGD:training_settings}.
%
As shown in Tab.~\ref{tab:urd:SOTA-PascalContext}, our HFGD achieved 63.8\% mIOU with single-scale without flipping and 64.9\% mIOU with multi-scales with flipping, outperforming previous state-of-the-art by 1\% mIOU in ECCV-2022.
HFGD is now the new state-of-the-art method on Pascal Context for the methods that only use the ImageNet pre-trained backbone without extra techniques~\cite{cAugReg,cFocalLoss,cVNet}.

\begin{figure*}[th]
\centering
\includegraphics[width=1.0\linewidth]{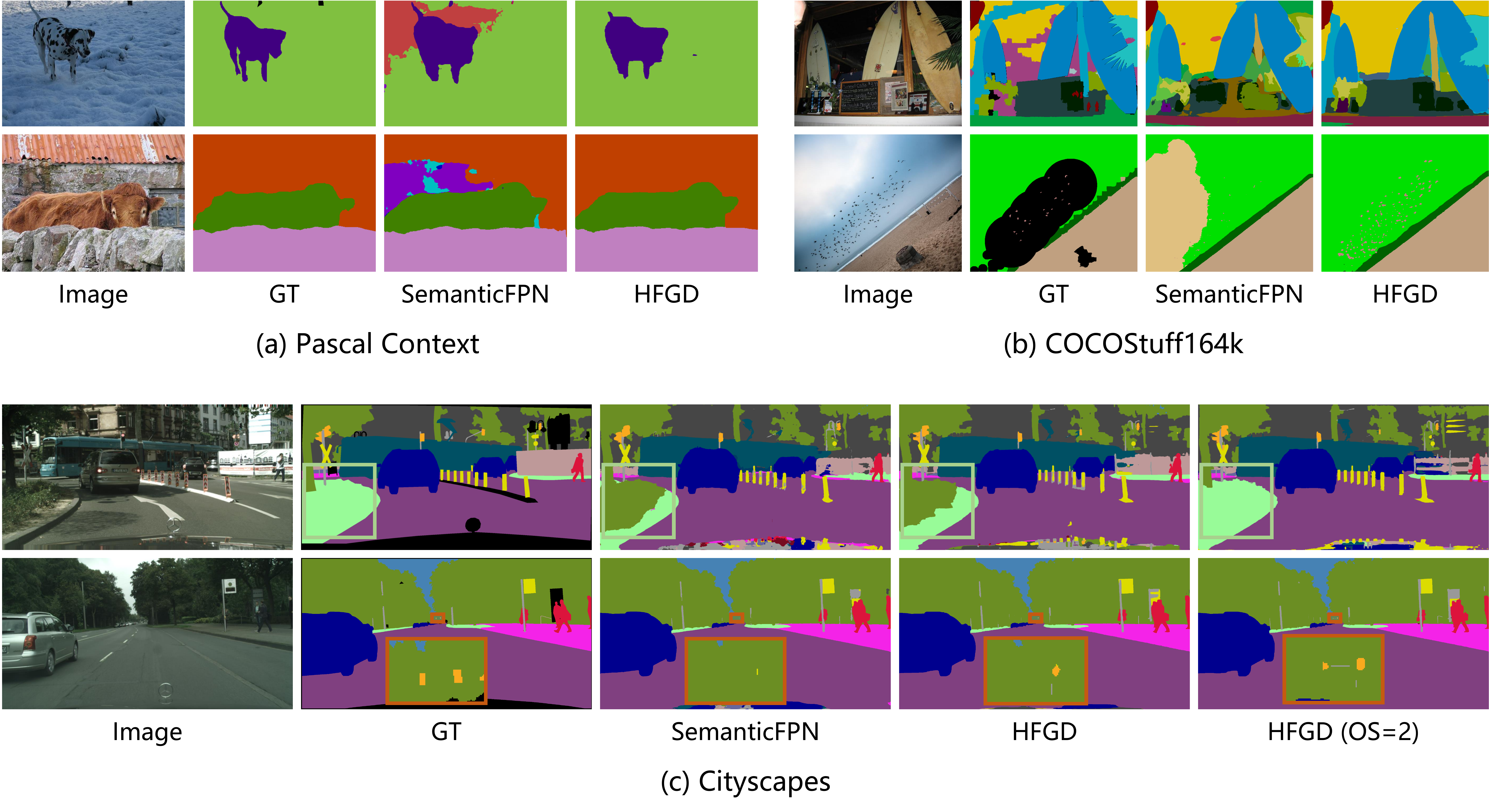}
\caption{
Visual comparisons 
between SemanticFPN ($OS=4$), HFGD ($OS=4$), and HFGD ($OS=2$).
Zoom in to see better.
The results are obtained using single-scale without flipping.
More visualizations are presented in the supplementary.
}
\label{fig:CFGD:main-R50-Vis}
\end{figure*}

\begin{figure}[th]
\centering
\includegraphics[width=1.0\linewidth]{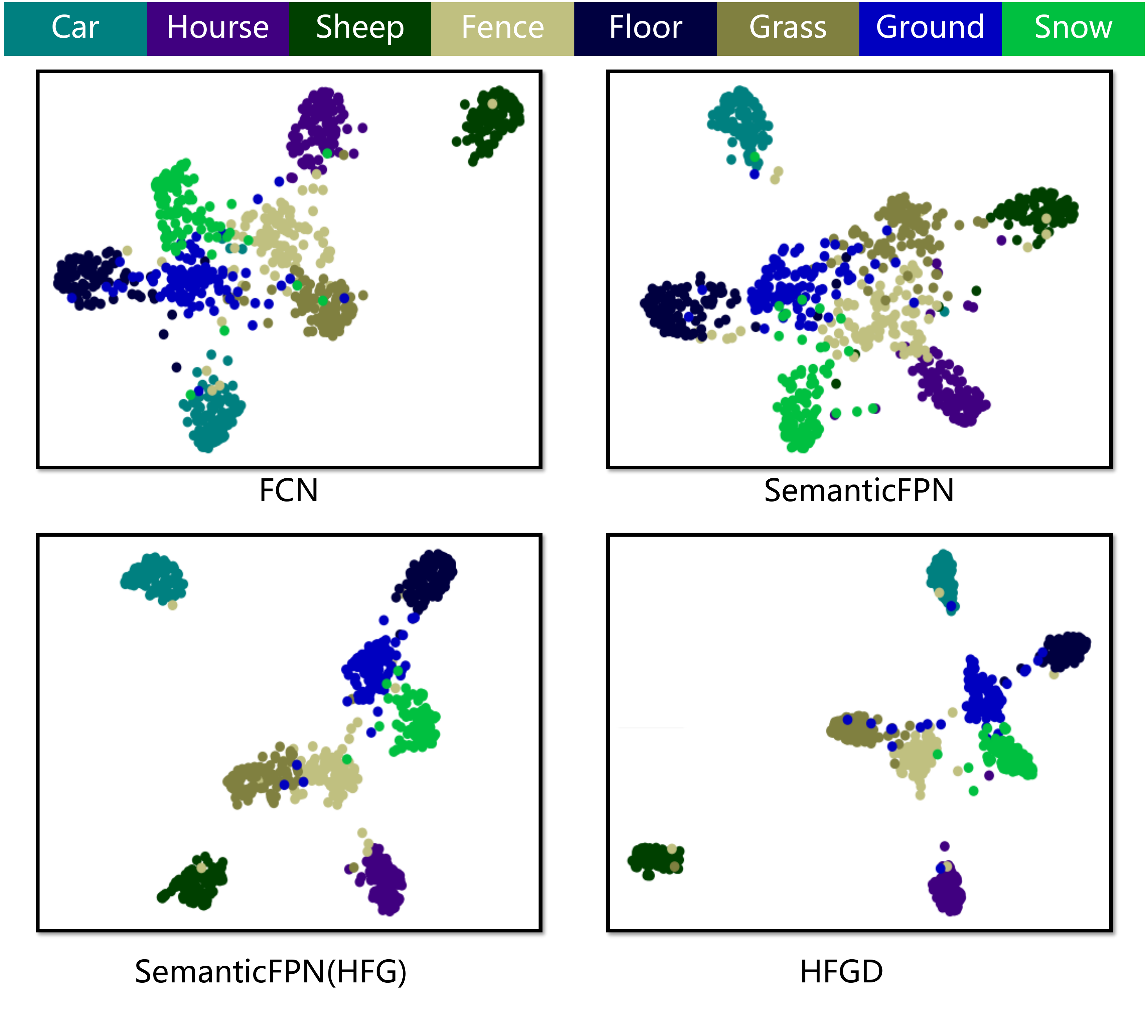}
\caption{
UMAP~\cite{cUMAP} Visualization on Pascal Context. 
HFGD features are most separable from the inter-class perspective and most compact from the intra-class perspective, resulting in the best accuracy (Tab.~\ref{tab:ablation_hfgd}).
}
\label{fig:umap_vis}
\end{figure}

\subsection{Apply high-level guide to existing upsamplers }

Though we recommend using U-SFPN + CAE to fit our proposed HFGM for efficiency, many existing methods attempt to improve the accuracy of upsamplers by improving intermediate-specific operations. 
For example, FaPN~\cite{cFaPN} uses SENet~\cite{cSENet} and deformable convolutions to try to align low-level features with high-level features.
UperNet~\cite{cUper} uses PPM~\cite{cPSPNet} to improve the high-level features of the backbone network.

From another perspective, HFGM directly optimizes the final upsampling quality rather than the intermediate process.
Thus, it should be able to improve the accuracy
of these upsamplers further, using high-level features as teachers to
constrain their upsampling results.

Using the ablation experimental setup in the main paper, Tab~\ref{tab:ablation_hfg_other_upsamplers} verifies our HFGM can effectively boost mIOU for different upsamplers. 
We believe that HFGM has generalizability to other similar upsamplers.
\section{Experiments on COCOStuff164k Dataset}

COCOStuff164k~\cite{cCocoStuff}, which has become popular in recent years, poses a great challenge for semantic segmentation models due to its high diversity (118k training images and 5000 testing images) and complexity (171 classes). 
We adopt ConvNeXt-Large as our backbone network and follow the training settings described in Sec.~\ref{sec:HFGD:training_settings} and we train our model for 40K iterations.
In Tab.~\ref{tab:HFGD:SOTA-COCOStuff164k}, we compare our proposed HFGD with other state-of-the-art methods.
HFGD outperforms the previous state-of-the-art~\cite{cSegNeXt} by a large margin (49.4\% vs 47.3\% mIOU).

\begin{table}[t]
\centering
\small
\resizebox{\linewidth}{!}
{\def\arraystretch{1} \tabcolsep=0.55em 
\begin{tabular}{l|c|c|c|c}
\toprule
Methods & Backbone & Avenue &\multicolumn{2}{c}{mIOU(\%)} \\
& & & SS & MF \\
\midrule
\midrule
SETR~\cite{cSETR}           & ViT-L           & CVPR'21 & - & 55.8 \\
DPT~\cite{cDPT}             & ViT-Hybrid      & ICCV'21 & - & 60.5 \\
Segmenter~\cite{cSegmenter} & ViT-L           & ICCV'21 & - & 59.0 \\
OCNet~\cite{cOCNet}         & HRNet-W48       & IJCV'21 & - & 56.2 \\
CAA~\cite{cCAA}             & EfficientNet-B7 & AAAI'22 & - & 60.5 \\
CAA + CAR~\cite{cCAR}        & ConvNeXt-L      & ECCV'22 & 62.7 & 63.9 \\
SegNeXt~\cite{cSegNeXt}     & MSCAN-L         & NIPS'22 & 59.2 & 60.9 \\
SegViT~\cite{cSegViT}*       & ViT-L            & NIPS'22 & - & 65.3 \\
TSG~\cite{cTSG}             & Swin-L           & CVPR'23 & - & 63.3 \\
IDRNet~\cite{cIDRNet}        & Swin-L           & NIPS'23 & - & 64.5 \\
\midrule
HFGD (OS=4)          & ConvNeXt-L & - & \textbf{63.8} & \textbf{64.9} \\
HFGD (OS=4)*          & ConvNeXt-L & - & \textbf{64.9} & \textbf{65.6}  \\
\bottomrule
\end{tabular}
}
\caption{
Comparisons to state-of-the-art methods on Pascal Context dataset.
%
\textit{SS}: Single-scale performance w/o flipping.
\textit{MF}: Multi-scale performance w/ flipping.
\textit{*}: Using stronger settings~(\eg{~\cite{cAugReg,cFocalLoss,cVNet}}) than Sec.~\ref{sec:HFGD:training_settings}.
``-'' in column \textit{SS} indicates that this result was not reported in the original paper.
}
\label{tab:urd:SOTA-PascalContext}
\end{table}

\begin{table}[t]
\centering
\small
\resizebox{\linewidth}{!}
{\def\arraystretch{1} \tabcolsep=0.55em 
\begin{tabular}{l|c|c|c|c}
\toprule
Methods & Backbone  & Avenue &\multicolumn{2}{c}{mIOU(\%)}\\
& & & SS & MF \\
\midrule
\midrule
OCR~\cite{cOCR,cHRFormer} & HRFormer-B & NIPS'21 & - & 43.3 \\
SegFormer~\cite{cSegFormer} & MiT-B5 & NIPS'21 & - & 46.7 \\
CAA~\cite{cCAA} & EfficientNet-B5 & AAAI'22 & - & 47.3 \\
SegNeXt~\cite{cSegNeXt} & MSCAN-L & NIPS'22 & 46.5 & 47.2 \\
RankSeg~\cite{cRankSeg} & ViT-L & ECCV'22 & 46.7 & 47.9 \\
\midrule
HFGD (OS=4) & ConvNeXt-L & - & \textbf{49.0} & \textbf{49.4}    \\ 
\bottomrule
\end{tabular}
}
\caption{
Comparisons to state-of-the-art methods on COCOStuff164k dataset.
%
%
%
\textit{SS}: Single-scale performance w/o flipping.
\textit{MF}: Multi-scale performance w/ flipping.
%
``-'' in column \textit{SS} indicates that this result was not reported in the original paper.
}
\label{tab:HFGD:SOTA-COCOStuff164k}
\end{table}
\section{Experiments on Cityscapes Dataset}

Cityscapes~\cite{cCityScapes} is a semantic segmentation dataset that consists of high-resolution images of road scenes with accurate annotations.
It has 19 labeled classes and contains 2975/500/1525 training/validation/test images.

\subsection{Ablation studies on feature map resolution}
\label{sec:exps:ablation_os2}
\begin{table}[t]
\centering
\small
\resizebox{\linewidth}{!}
{\def\arraystretch{1} \tabcolsep=1.15em 
\begin{tabular}{l|c|c|l} 
\toprule
Method & Backbone & $OS$ & mIOU(\%)  \\
\midrule
\midrule
SemanticFPN & ResNet-50 & 4      & 76.44   \\
\midrule
HFGD & ResNet-50  & 4      & 79.11    \\
HFGD & ResNet-50  & 2 & \textbf{79.81} (\textcolor{black}{+0.7}) \\
\bottomrule
\end{tabular}
}
\caption{
Ablation studies on different output stride settings in U-SFPN on Cityscapes dataset.
}
\label{tab:ablation_ct_os2}
\end{table}

We mainly conduct ablation experiments on Cityscapes to verify the superiority of using ultra high resolution feature maps (i.e. $OS=2$) since its GT annotations are the most accurate.
We use ResNet-50 as the backbone and train SemanticFPN, HFGD ($OS=4$), and HFGD ($OS=2$) for 30K iterations following the training settings in Sec.~\ref{sec:HFGD:training_settings}. 
As shown in Tab.~\ref{tab:ablation_ct_os2}, the $OS=2$ has further improved the accuracy over $OS=4$ by 0.7\% mIOU (79.11 vs 79.81).

\subsection{Comparison with the state-of-the-art methods}
To compare our method against the state-of-the-art, we use ConvNeXt-Large and follow the training settings described in Sec.~\ref{sec:HFGD:training_settings}.
Note that we only compare with methods that are trained only on the Cityscapes fine annotations, similar to many works~\cite{cSegFormer,cKMaXDeepLab}.
We set the crop size to 513$\times$1025~\cite{cDeepLabV3,cPanopticDeepLab} and train our HFGD model for 60K iterations for the $OS=4$ version and 80K iterations for the $OS=2$ version.
As shown in Tab.~\ref{tab:HFGD:SOTA-Cityscapes}, the proposed HFGD outperforms the previous state-of-the-art proposed in recent years on the Cityscapes validation set.

A similar comparison is conducted on the Cityscapes test set, where we set batch size = 32 following kMaXDeepLab~\cite{cKMaXDeepLab}
while the other settings remain unchanged.
Results on the test set can fairly demonstrate the effectiveness of the proposed method since no ground-truths are provided.
Note that we did not use hard sample mining.
As shown in Tab.~\ref{tab:HFGD:test-Cityscapes}, our HFGD sets new state-of-the-art on Cityscapes test set (when only using fine set for training).

\begin{table}[t]
\centering
\small
\resizebox{\linewidth}{!}{
\begin{tabular}{l|c|c|c|c}
\toprule
Methods &Backbone & Avenue &\multicolumn{2}{c}{mIOU(\%)}\\
&  & & SS & MF \\
\midrule
\midrule
RepVGG\cite{cRepVGG} & RepVGG-B2 & CVPR'21 & - & 80.6 \\
SETR~\cite{cSegFormer} &ViT-L & CVPR'21 & - & 82.2 \\
Segmenter~\cite{cSegmenter} &ViT-L & ICCV'21 & - & 81.3 \\
OCR~\cite{cOCR,cHRFormer} &HRFormer-B & NIPS'21 & - & 82.6 \\
HRViT-b3~\cite{cHRViT}  & MiT-B3 & CVPR'22 & - & 83.2\\
FAN-L~\cite{cFANs} & FAN-Hybrid & ICML'22 & - & 82.3 \\
SegDeformer~\cite{cSegDeformer} & Swin-L & ECCV'22 & - & 83.5 \\
GSS-FT-W~\cite{cGSS} & Swin-L & CVPR'23 & - & 80.5 \\
TSG~\cite{cTSG} & Swin-L & CVPR'23 & - & 83.1 \\
STL~\cite{cSTL} & FAN-Hybrid & ICCV'23 & - & 82.8 \\
DDP(Step 1)~\cite{cDDP} & ConvNeXt-L & ICCV'23 & 83.0 & 83.8 \\
DDP(Step 3)~\cite{cDDP} & ConvNeXt-L & ICCV'23 & 83.2 & 83.9 \\
\midrule
HFGD (OS=4)     & ConvNeXt-L         & - & 83.1 & 83.8 \\
HFGD (OS=2)    & ConvNeXt-L              & - & 83.2 & \textbf{84.0} \\
\bottomrule
\end{tabular}
}
\caption{
Comparisons to state-of-the-art methods on Cityscapes validation set.
%
\textit{SS}: Single scale performance w/o flipping.
\textit{MF}: Multi-scale performance w/ flipping.
``-'' in column \textit{SS} indicates that this result was not reported in the original paper.
}
\label{tab:HFGD:SOTA-Cityscapes}
\end{table}

\begin{table}[th!]
\centering
\small
\resizebox{\linewidth}{!}{
\begin{tabular}{l|c|c|c}
\toprule
Methods &Backbone & Avenue &mIOU(\%)\\
\midrule
\midrule
Panoptic-DeepLab\cite{cPanopticDeepLab} & SWideRNet & CVPR'20 & 80.4 \\
Axial-DeepLab\cite{cAxialDeepLab} & Axial-ResNet-XL & ECCV'20 & 79.9 \\
SETR\cite{cSETR} & ViT-L & CVPR'21 & 81.1 \\
SegFormer\cite{cSegFormer} & MiT-B5 & NIPS'21 & 82.2 \\
kMaXDeepLab\cite{cKMaXDeepLab} (OS=2) & ConvNeXt-L & ECCV'22 & 83.2 \\
\midrule
HFGD (OS=2)    & ConvNeXt-L & -  & \textbf{83.3} \\
\bottomrule
\end{tabular}
}
\caption{
Comparisons to state-of-the-art methods on Cityscapes test set.
We only list methods trained on Cityscapes fine annotation set for fair comparisons.
Due to their policy, we are only able to present a single result.
}

\label{tab:HFGD:test-Cityscapes}
\end{table}

\section{Visualizations}
We present visual comparisons between SemanticFPN and our HFGD on Pascal Context, CocoStuff and Cityscapes in Fig~\ref{fig:CFGD:main-R50-Vis}.
HFGD is clearly more capable of segmenting small/thin objects. 
For example, HFGD even performs better than manual annotation when segmenting very small birds in the sky.
When using OS=2 on Cityscapes, HFGD can segment the remote traffic light partially missed in OS=4
which is of great importance in autonomous driving.

\vspace{1mm}
\noindent\textbf{UMAP:} We conducted UMAP~\cite{cUMAP} visualization in Fig.~\ref{fig:umap_vis} to analyze the impact of our HFG/HFGD in addressing the issue presented in Fig.~\ref{fig:hfgd:concept}. 
The final layer features used for classification are used (the features produced by the upsampler for HFGD) in this visualization.
Since Pascal Context has 59 classes, we only selected 8 representative error-prone classes (\ie, hard to discriminate, \eg Floor vs Grass vs Ground vs Snow) in Fig.~\ref{fig:umap_vis} for clarity.
%
The class discrimination ability of SemanticFPN is slightly worse than FPN.
``SemanticFPN (HFG)'' restores this ability with the help of HFG.
Additionally, our entire HFGD results in more discriminative inter-class and more compact intra-class features.
Our full solution HFGD obtains more discriminative features (\ie more separable inter-class distance and more compact intra-class representation).

%
\section{Conclusion }
In this paper, we propose to use the high-level features as the teacher to guide the training of the upsampler branch (student), resulting in an effective and efficient decoder framework.
Specifically, the core of our method is using high-level features as guidance (HFG) and proper stop-gradient operations for the upsampler learning, which effectively addresses the observed issues in Fig.~\ref{fig:hfgd:concept}.
In addition, we explore a context-augmented encoder (CAE) to effectively enhance the OS=32 high-level features
and propose a modified version of SemanticFPN to better fit our HFGM.
With thorough experiments, HFGD achieves largely improved performance over previous upsampling-based state-of-the-art methods that does not use extra training data, on Pascal Context, COCOStuff, and Cityscapes.
HFGD also achieves slightly better results than dilation-based CNNs but using much less computation cost (\eg Self-Attention + CAR~\cite{cCAR}).
%
We refer the readers to our appendix for more ablation studies, experiments, implementation details, and visualizations.

\clearpage

\clearpage

\appendix

{
   \newpage
       \twocolumn[
        \centering
        \Large
        \vspace{0.5em}\textbf{Appendix}\\
        \vspace{2.5em}
       ] 
   }


\section{Extra details}

We show the detailed setting of the Ultra SemanticFPN (U-SFPN) here.
U-SFPN is based on SemanticFPN~\cite{cPanopticFPN} with some modifications to better cooperate with high-level feature guidance. 
Note that the right side of SemanitcFPN was originally 128-D. 
For fairness, all our experiments are set to 256-d (mentioned in the main paper).

\begin{figure}[th]
\centering
\includegraphics[width=1.0\linewidth]{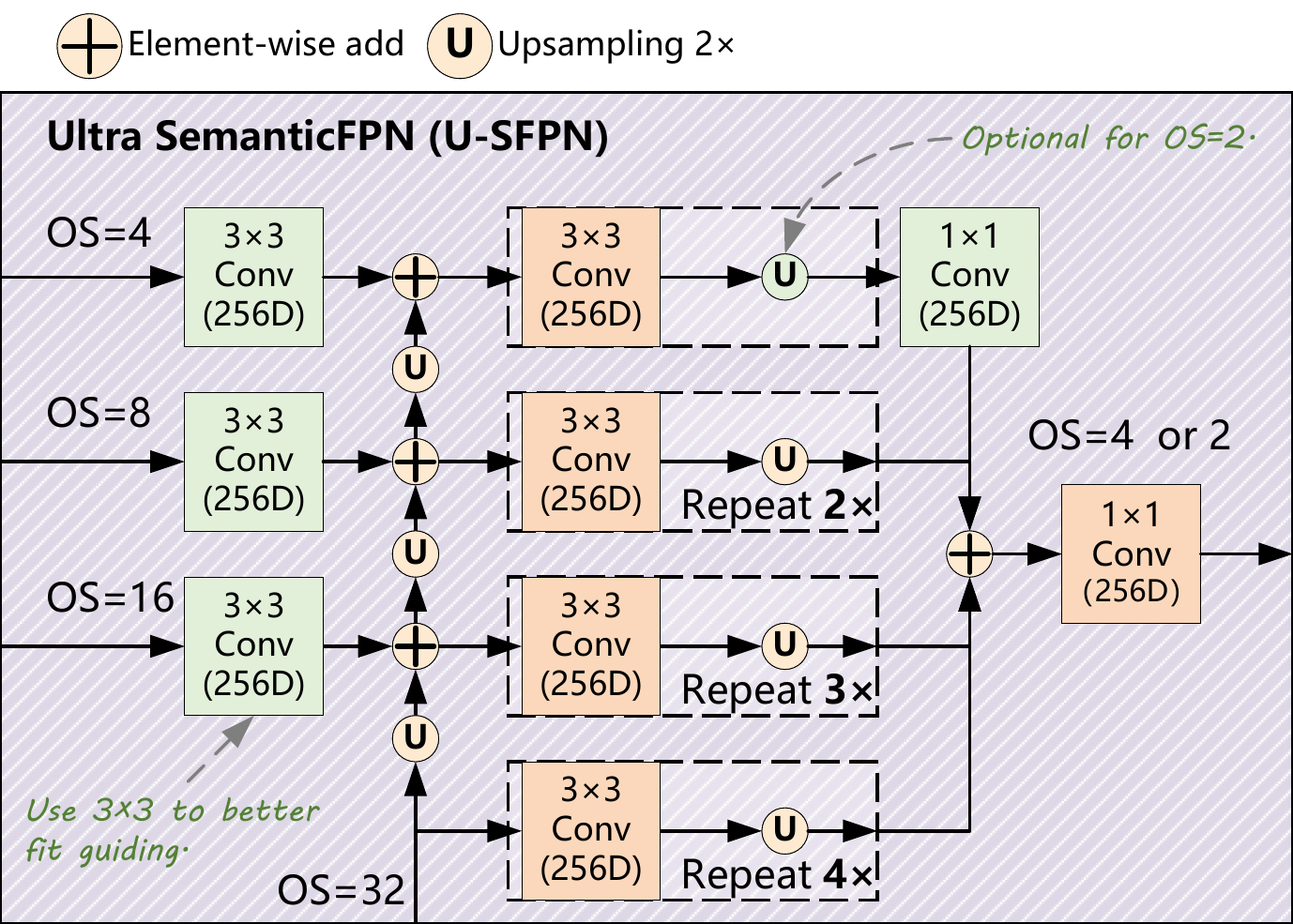}
\caption{
Architecture of U-SFPN.
Ultra SemanticFPN (U-SFPN) is modified from SemanitcFPN (SFPN)~\cite{cPanopticFPN} to better cooperate with high-level feature guidance. 
We highlight the modified layers/operations with a green background.
}
\label{fig:CFGD:supp:vis:USFPN}
\end{figure}

\section{Extra experiments}

\subsection{Experiments on ADE20k}

In addition to the datasets mentioned in the main text, we also compared our HFGD on ADE20k. 
ADE20k dataset comprises of 150 classes and is divided into 20210/2000 for training and validation purposes.
Following other works~\cite{cKMaXDeepLab,cDDP},
we use $641\times641$ crop size, and other settings are the same as the main paper. 
The results are shown in Tab.~\ref{tab:HFGD:supp:SOTA-ADE20K}.
HFGD significantly outperforms the other methods.

\begin{table}[th]
\centering
\small
\resizebox{\linewidth}{!}
{
\begin{tabular}{l|c|c|c|c}
\toprule
Methods & Backbone  & Avenue &\multicolumn{2}{c}{mIOU(\%)}\\
& & & SS & MF \\
\midrule
\midrule
SETR~\cite{cSETR} & ViT-L & CVPR'21 & - & 50.0 \\
Segmenter~\cite{cSegmenter} & ViT-L & ICCV'21 & 51.7 & 53.6 \\
SegFormer~\cite{cSegFormer} & MiT-B5 & NIPS'21 & 51.0 & 51.8 \\
Uper~\cite{cUper,cConvNeXT} & ConvNeXt-L & CVPR'22 & - & 53.7 \\
SegNeXt-L~\cite{cSegNeXt} & MSCAN-L & NIPS'22 & 51.0 & 52.1 \\
TSG~\cite{cTSG} & Swin-L & CVPR'23 & - & 54.2 \\
DDP~\cite{cDDP}(step1) & Swin-L & ICCV'23 & 53.1 & 54.4   \\
DDP~\cite{cDDP}(step3) & Swin-L & ICCV'23 & 53.2 & 54.4  \\
\midrule
HFGD & ConvNeXt-L & - & \textbf{54.3} & \textbf{55.4}    \\ 
\bottomrule
\end{tabular}
}
\caption{
Comparisons to state-of-the-art methods on ADE20k dataset.
\textit{SS:} Single-scale performance w/o flipping.
\textit{MF:} Multi-scale performance w/ flipping.
``-'' in column \textit{SS} indicates that this result was not reported in the original paper.
}
\label{tab:HFGD:supp:SOTA-ADE20K}
\end{table}

\subsection{Experiments on COCOStuff10k}

We also test our HFGD on COCOStuff10k, which is a subset of COCOStuff164k. 
The subset is split into 9000/1000 for training and testing.
Our training settings are identical to COCOStuff164k from the main paper.
The results are shown in Tab.~\ref{tab:HFGD:supp:SOTA-COCOSTUFF10K}. HFGD significantly outperforms the other methods.

\begin{table}[th]
\centering
\small
\resizebox{\linewidth}{!}
{
\begin{tabular}{l|c|c|c|c}
\toprule
Methods & Backbone  & Avenue &\multicolumn{2}{c}{mIOU(\%)}\\
& & & SS & MF \\
\midrule
\midrule
OCR~\cite{cOCR} & HRNet-W48~\cite{cHRNet} & ECCV'20 & - & 45.2 \\
CAA~\cite{cEfficientNet} & EfficientNet-B7~\cite{cEfficientNet} & AAAI'22 & - & 45.4\\
RankSeg~\cite{cRankSeg} & ViT-L & ECCV'22 & - & 47.9 \\
CAA + CAR~\cite{cCAR} & ConvNeXt-L & ECCV'22 & 49.0 & 50.0 \\
SegNeXt-L~\cite{cSegNeXt} & MSCAN-L & NIPS'22 & 46.4 & 47.2 \\
\midrule
HFGD & ConvNeXt-L & - & \textbf{49.5} & \textbf{50.3}    \\ 
\bottomrule
\end{tabular}
}
\caption{
Comparisons to state-of-the-art methods on COCOStuff10k dataset.
\textit{SS:} Single-scale performance w/o flipping.
\textit{MF:} Multi-scale performance w/ flipping.
``-'' in column \textit{SS} indicates that this result was not reported in the original paper.
}
\label{tab:HFGD:supp:SOTA-COCOSTUFF10K}
\end{table}

\subsection{Advantage of pre-training on COCOStuff10k}

In addition to Pascal Context (Tab.~\ref{tab:ablation_imagenet_simple} in the main paper),
We also test its advantage on COCOStuff10k, which contains 9000 training images (only 4998 in the Pascal Context) and approaches the 10K COCO training images mentioned in~\cite{cRethinkingImageNet}.
%
%
We find that despite the increase in training samples (Tab.~\ref{tab:supp:ablation_imagenet_simple_coco}), the huge advantage of ImageNet pre-training remains undiminished. 

In fact, we are not the first to be aware of this issue. 
As early as 2019, NAS-based AutoDeepLab~\cite{cAutoDeepLab}, although using the COCO dataset (which can be seen as COCOStuff164k) for pre-training, due to the lack of ImageNet pre-training, it still has slightly worse mIOU compared to other methods.
%

\begin{table}[th]
\centering
\small
\resizebox{\linewidth}{!}
{\def\arraystretch{1} \tabcolsep=0.6em 
\begin{tabular}{l|c|c|c} 
\toprule
Training Iterations & 30K & 90K & 180K  \\
\midrule
\midrule
ResNet-50 (ImageNet) + FCN & 32.92 & - & - \\
ResNet-50 (scratch) + FCN  & 16.70 & 22.62 & 24.62 \\
\bottomrule
\end{tabular}
}
\caption{Simple experiments to present the importance and advantage of ImageNet pre-train.
Trained on COCOStuff10k dataset.
Results are in mIOU(\%).
}
\label{tab:supp:ablation_imagenet_simple_coco}
\end{table}

\begin{figure*}[ht]
\centering
\includegraphics[width=1.0\linewidth]{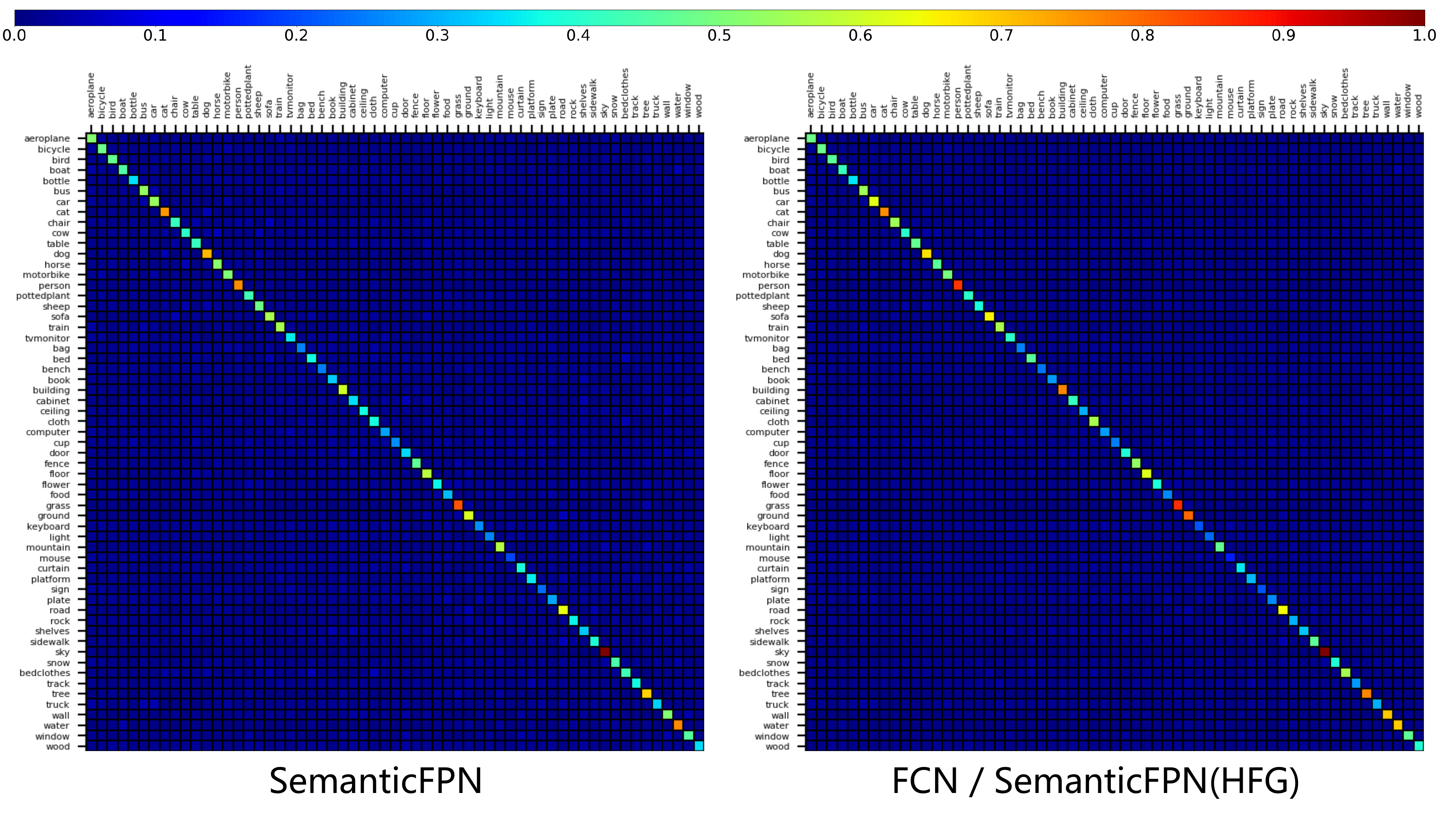}
\caption{
Visualization of inter-class dependencies among all the class tokens for SemanticFPN and FCN/SemanticFPN (HFG).
Zoom in to see better (\eg class names).
More high-score (\eg closer to the red) in the diagonal indicates better class discriminability.
SemanticFPN relies more on inter-class representation than FCN/SemanticHFG for classification.
See section~\ref{sec:supp:vis-class-token} for details.
}
\label{fig:CFGD:supp:vis:class_token}
\end{figure*}

\section{Extra visualizations}

\subsection{Visualization on class token}
\label{sec:supp:vis-class-token}

In this section, we visualize the inter-class dependencies between class tokens of ``SemanticFPN'' and ``FCN/SemanitcFPN (HFG)'' (When using HFG, SemanticFPN uses the same token as FCN).
Our visualization approach is inspired by the CAR~\cite{cCAR}, where the major difference is that CAR visualizes the class centers extracted from the feature map, while we visualize the class token for final classification.
According to Fig.~\ref{fig:CFGD:supp:vis:class_token}, SemanticFPN's class tokens have slightly lower diagonal confidence scores than FCN (around 20\%), which means SemanticFPN relies more on other classes' information during classification.
. 
%
By using the proposed HFG, the class token's discriminability in SemanticFPN is no longer weakened. 
This results in high-resolution and robust class representation at the same time. (Please refer to the main paper's experiments section for further details).

\subsection{Visual results on Pascal Context dataset}

\begin{figure*}[t]
\centering
\includegraphics[width=0.8\linewidth]{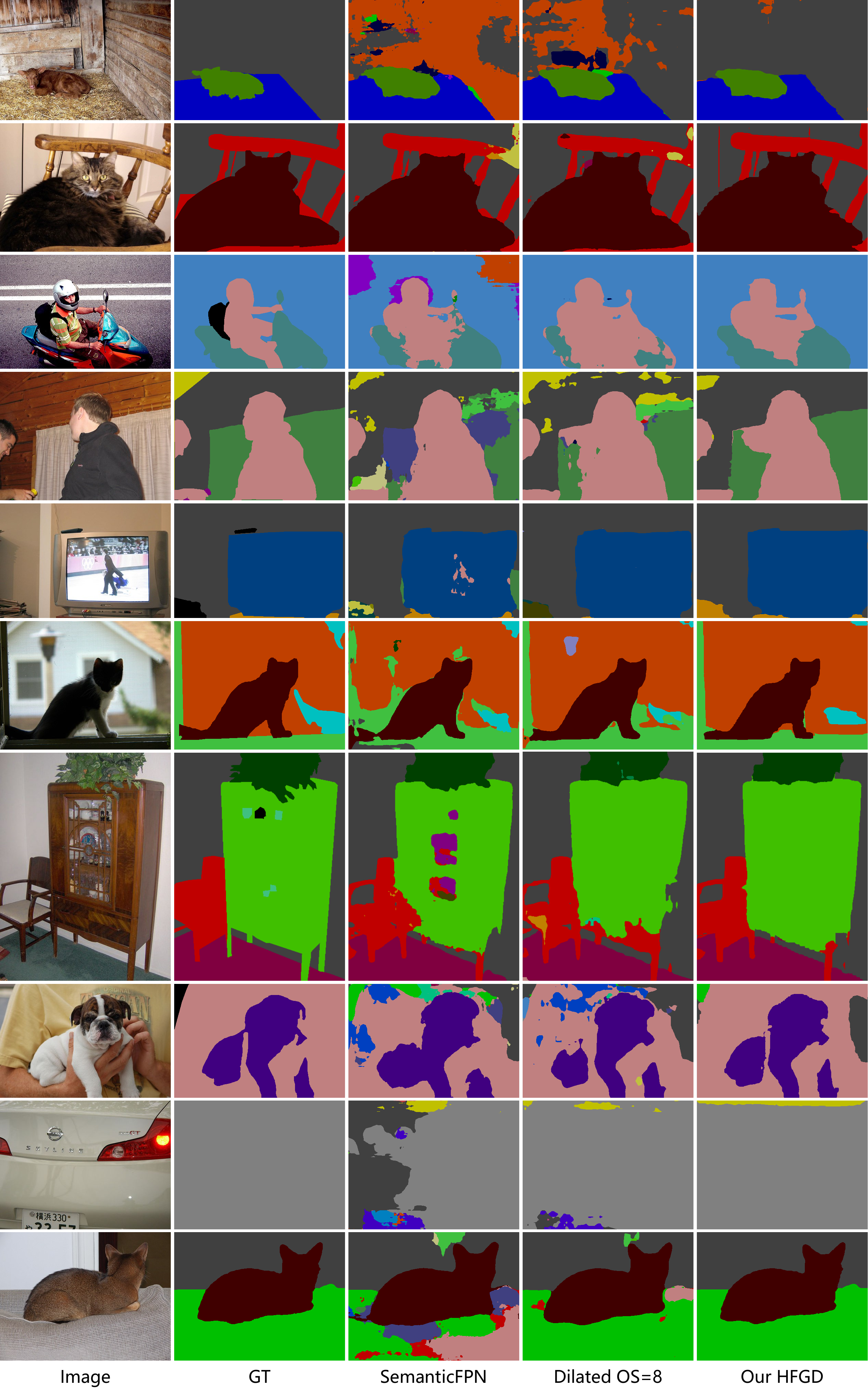}
\caption{
Visual comparisons between SemanticFPN ($OS=4$), Self-Attention(CAR)~($OS=8$), and HFGD ($OS=4$) on the Pascal Context dataset.
Zoom in to see better.
The results are obtained using single-scale without flipping.
}
\label{fig:HFGD:supp:vis:PascalContext}
\end{figure*}

For Pascal Context, we compare visual results
of SemanticFPN, Self-Attention (CAR~\cite{cCAR})~(Dilation $OS=8$), and our HFGD in Fig.~\ref{fig:HFGD:supp:vis:PascalContext}. 
We can see that SemnaitcFPN has better details than SA while
SA has better class discrimination than SemanitcFPN. 
Our HFGD performs the best in both perspectives.

\subsection{Visual results on COCOStuff164k dataset}

\begin{figure*}[th]
\centering
\includegraphics[width=0.6\linewidth]{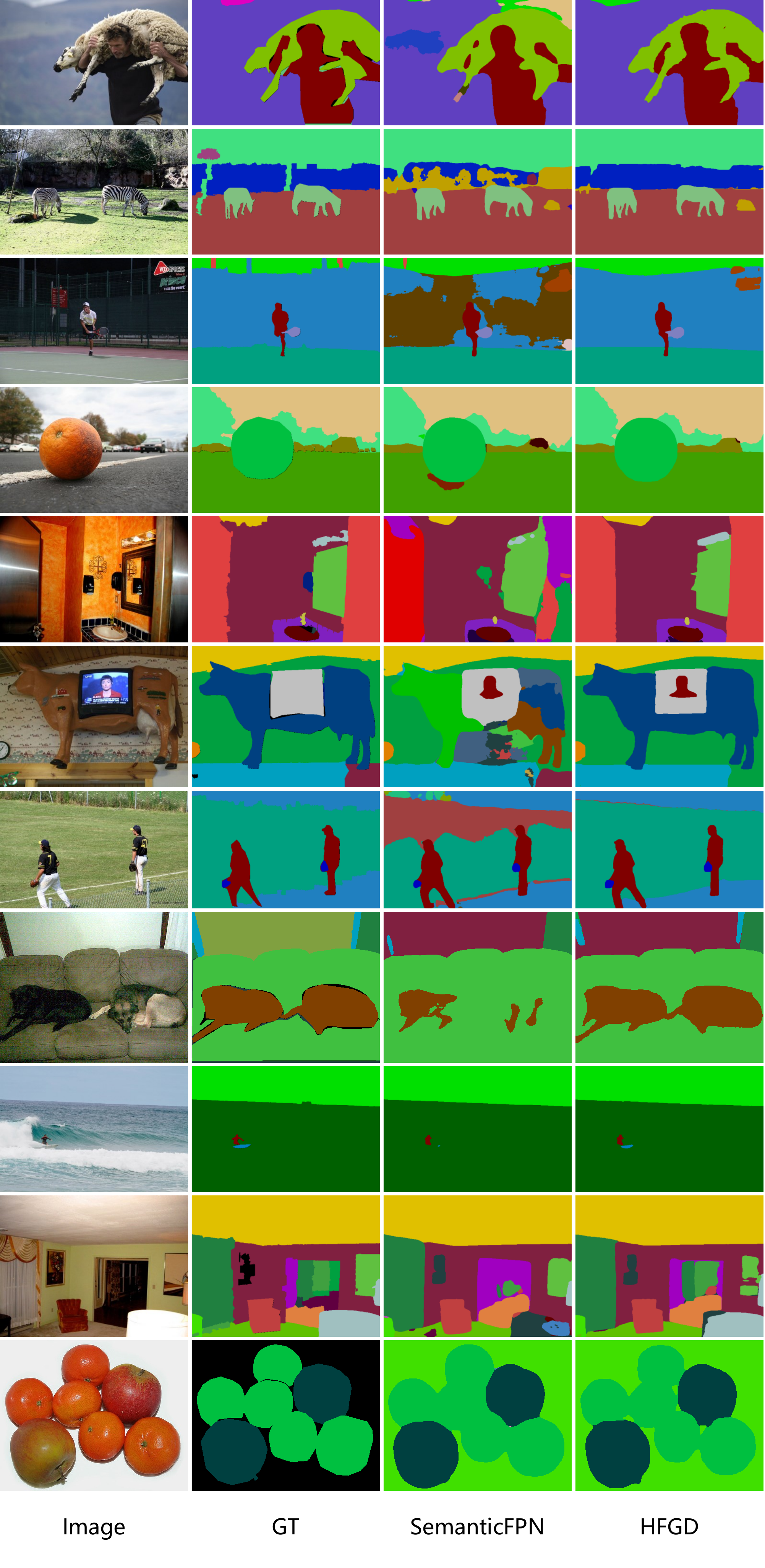}
\caption{
Visual comparisons between SemanticFPN ($OS=4$) and HFGD ($OS=4$) on the COCOStuff164k dataset.
Zoom in to see better.
The results are obtained using single-scale without flipping setting.
}
\label{fig:HFGD:supp:vis:COCOStuff}
\end{figure*}

We then compare COCOStuff164k results of SemanticFPN and our HFGD in Fig.~\ref{fig:HFGD:supp:vis:COCOStuff}. 
We observed that our HFGD provides better class discrimination and finer details than SemanticFPN for scenes, both indoor and outdoor, at close-up and long-range distances.

\subsection{Visual results on Cityscapes dataset}

\begin{figure*}[th]
\centering
\includegraphics[width=\linewidth]{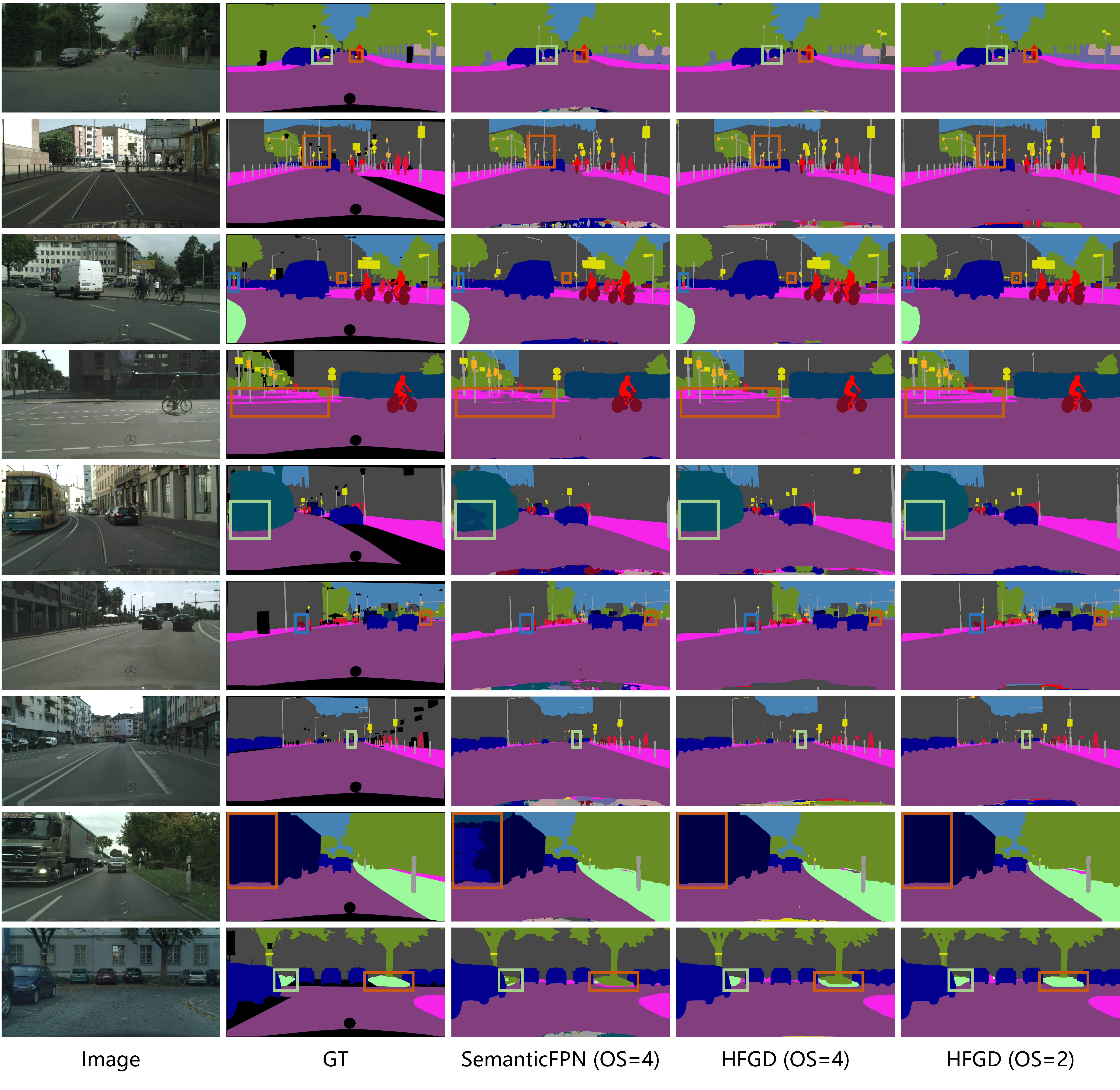}
\caption{
Visual comparisons between SemanticFPN ($OS=4$), HFGD ($OS=4$), and HFGD ($OS=2$) on the Cityscapes dataset.
Zoom in to see better.
The results are obtained using single-scale without flipping.
}
\label{fig:HFGD:supp:vis:Cityscapes}
\end{figure*}

Finally, we compare the visualizations of SemanticFPN and our HFGD on the Cityscapes dataset in Fig.~\ref{fig:HFGD:supp:vis:Cityscapes}.
Our proposed HFGD outperforms SemanticFPN, particularly for small objects that are far away.
such as street signs and signal lights. 
This is critical to safe autonomous driving perception.

\clearpage
{
    \small
    \bibliographystyle{ieeenat_fullname}
    \bibliography{main}
}


\end{document}